\documentclass[runningheads]{llncs}

\usepackage{eccv}

\usepackage{eccvabbrv}
\usepackage[inkscapelatex=false]{svg}
\usepackage{graphicx}
\usepackage{booktabs}
\usepackage{multirow}
\usepackage[accsupp]{axessibility}  % Improves PDF readability for those with disabilities.
\usepackage{booktabs}
\usepackage{mathrsfs}
\usepackage{marvosym}
\newcommand{\tbl}[1]{Tab. \ref{#1}}
\newcommand{\fig}[1]{Fig. \ref{#1}}
\newcommand{\eq}[1]{Eq. (\ref{#1})}

\usepackage[pagebackref,breaklinks,colorlinks,citecolor=eccvblue]{hyperref}
\usepackage{orcidlink}

\begin{document}

\title{PromptIQA: Boosting the Performance and Generalization for No-Reference Image Quality Assessment via Prompts} 


\author{Zewen Chen\inst{1,2}$^{\star}$\orcidlink{0000-0001-7791-0959}
 \and
 Haina Qin\inst{1,2}\thanks{Equal contribution}\orcidlink{0009-0007-9477-2296} \and
Juan Wang\inst{1}\orcidlink{0000-0002-3848-9433}  \and
Chunfeng Yuan\inst{1}\orcidlink{0000-0003-2219-4961} \and
Bing  Li\inst{1}\textsuperscript{\Letter}\orcidlink{0000-0001-6114-1411} \and
Weiming Hu\inst{1,2,3} \orcidlink{0000-0001-9237-8825} \and
Liang Wang\inst{4} 
}
\authorrunning{Chen et al.}

\institute{State Key Laboratory of Multimodal Artificial Intelligence Systems, CASIA \and
School of Artificial Intelligence, University of Chinese Academy of Sciences \and
School of Information Science and Technology, ShanghaiTech University \and
OPPO \\
\email{\{chenzewen2022,qinhaina2020,jun\_wang\}@ia.ac.cn}, \email{\{cfyuan,bli,wmhu\}@nlpr.ia.ac.cn},\email{leon.wang@oppo.com}
}

\maketitle

\begin{abstract}
Due to the diversity of assessment requirements in various application scenarios for the IQA task, existing IQA methods struggle to directly adapt to these varied requirements after training. Thus, when facing new requirements, a typical approach is fine-tuning these models on datasets specifically created for those requirements. However, it is time-consuming to establish IQA datasets. In this work, we propose a Prompt-based IQA (PromptIQA) that can directly adapt to new requirements without fine-tuning after training. On one hand, it utilizes a short sequence of Image-Score Pairs (ISP) as prompts for targeted predictions, which significantly reduces the dependency on the data requirements. On the other hand, PromptIQA is trained on a mixed dataset with two proposed data augmentation strategies to learn diverse requirements, thus enabling it to effectively adapt to new requirements. Experiments indicate that the PromptIQA outperforms SOTA methods with higher performance and better generalization. The code will be available.
  \keywords{NR-IQA \and Image-Score Pairs \and Assessment Requirement}
\end{abstract}

\section{Introduction}

Image quality assessment (IQA) is a long-standing research in image processing fields. Compared to full-reference and reduced-reference IQA, no-reference IQA (NR-IQA) receives more attention since it removes the dependence on reference images which are even impossible to obtain in real-world applications.

IQA is a notably intricate and varied task, as the assessment requirements vary for different application scenarios. Thus, images with the same quality scores but sampled from different datasets may convey different subjective perceptions \cite{zhang2021uncertainty}.
This variety is more pronounced for the IQA datasets applied in different tasks, such as those for nature image IQA \cite{larson2010most, sheikh2006statistical}, underwater IQA \cite{yang2021reference}, and artificial intelligent (AI)-generated IQA \cite{li2023agiqa, wang2023aigciqa2023} \etc. In light of this, in the process of creating IQA datasets, it is crucial to clearly represent the assessment requirements to annotators for ensuring the accuracy and consistency of the quality score labels. 

\begin{figure}[tb]
    \centering
    \includegraphics[width=\columnwidth]{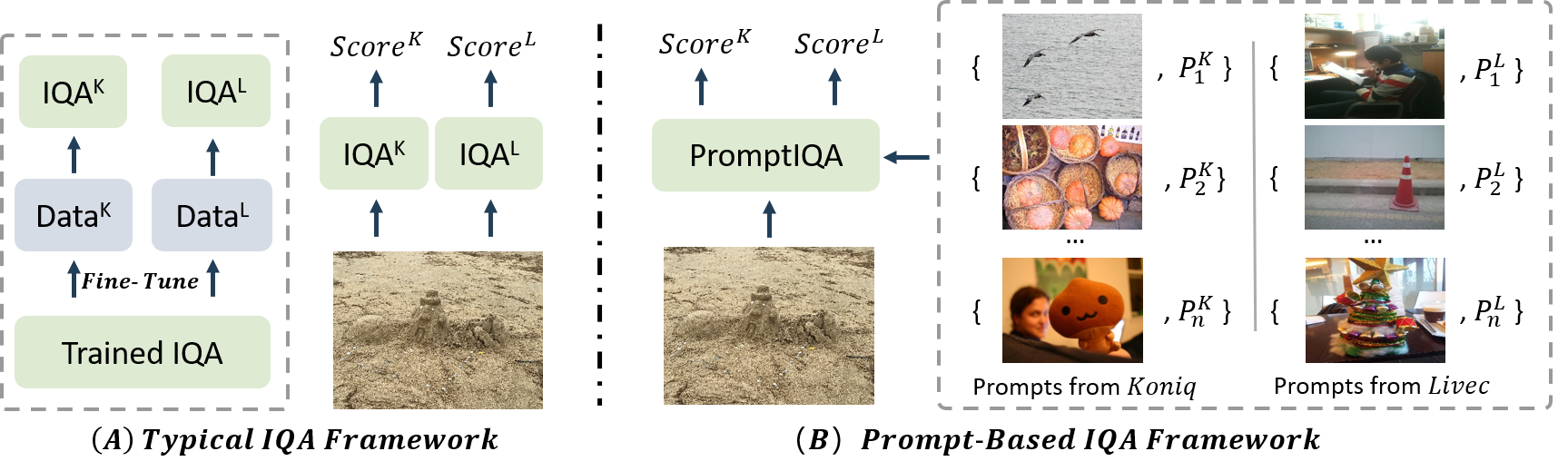}
    \caption{Comparison of the typical and prompt-based IQA frameworks. For a dataset $M$, pairs of images and corresponding scores $P_i^M$ constitute the ISP prompts, which represent the assessment requirement for the dataset $M$. 
    }
    \vspace{-0.7cm}
    \label{fig:idea}
\end{figure}

As shown in \fig{fig:idea} (A), most IQA models \cite{ke2021musiq, chen2023topiq, zhang2018blind, su2020blindly, yang2022maniqa, chen2024gmciqa} solely rely on the information contained in the input images to predict quality scores once the models are trained. Such an approach dose not take various assessment requirements into consideration, thereby potentially hampering the adaptability of these models to new assessment requirements. One direct approach is training or fine-tuning the models on datasets made by the new requirements. However, making IQA datasets is time-consuming and labor-intensive. Some methods try to reduce the data requirement for fine-tuning by proposing multi-dataset mixed training strategies \cite{zhang2021uncertainty, wang2021active} or carefully designing networks \cite{wang2023Hierarchical, qin2023data}. Nevertheless, in the real-world applications, the assessment requirements are usually various. Thus, it is essential to create new datasets aligned with these requirements and retrain or fine-tune IQA models on these datasets. The aforementioned methods still require considerable scale of datasets, making it impractical to construct datasets for each requirement. In this sight, it is more urgent to make IQA models effectively and low-costly comprehend assessment requirements with an exactly small number of data to better learn how to judge quality scores.

In this paper, we propose a novel NR-IQA framework that can effectively adapt to new assessment requirements with significantly fewer data and without  fine-tuning. In the process of creating dataset, organizers need to define the assessment requirement and select examples for this requirement to align annotators to the standards \cite{wu2023q}. Inspired by the fact that annotators must first comprehend the assessment requirement before assessing, we propose to teach the model to comprehend the assessment requirements through a kind of prompts, enabling adaptive score predictions on the targeted requirement. We term this method Prompt-based IQA (PromptIQA). As shown in \fig{fig:idea}, different from the typical IQA framework that have to be fine-tuned on the targeted datasets, the PromptIQA can pertinently predict scores by employing multiple image-score pairs (ISP) as the prompt without fine-tuning after the training is completed. 

One ISP prompt (ISPP), consisting of a specific number of ISPs extracted from one dataset, can represent the corresponding assessment requirement. There are two strategies to obtain the ISPPs: interval sampling and random sampling. Interval sampling selects ISPPs uniformly according to the score distribution of the dataset. In contrast, random sampling randomly selects ISPPs from the dataset, which is more similar to the process of making ISPPs for new requirements. In the experiment, we discuss the impact of these two strategies on the performance of PromptIQA.
Intuitively, the ISPPs extracted from different datasets can help the PromptIQA distinguish different requirements, which makes it is easy to achieve mixed training on a combination of datasets without intricate modifications.
The more assessment requirements the PromptIQA learns, the better adaption ability to unknown requirements it shows. In this work, we select a total of 12 datasets from 5 IQA tasks for mixed training. Specifically, in the training process, for each iteration, we randomly extract the ISPPs from the training set in one dataset, and the PromptIQA learns the assessment requirement from these ISPPs to predict scores for other images. When it comes to a new requirement, an ISPP aligned with the new requirement can directly teach the PromptIQA to learn the requirement, eliminating the need for extensive IQA datasets. 
Compared to text-based prompts that may exhibit bias \cite{wu2024qbench}, the ISPP that consist of multiple ISPs can more intuitively reflect assessment requirements, such as preferences, criteria, \etc 

In the training stage, the optimization objective of the PromptIQA is to minimize the discrepancy between the predicted scores and the ground truth (GT) labels. However, the GT labels in one dataset can only reflect one type of assessment requirement, which means the GT labels do not vary with changes in ISPPs. Hence, the model can find a shortcut to directly predict the quality score only depending on the input images instead of comprehending the assessment requirement through the ISPP first, which makes the ISPP ineffective. To solve this problem, we enrich the assessment requirements in one dataset by making a strong correlation between GT labels and the ISPPs, to ensure the PromptIQA can learn from the ISPPs. We propose two data augmentation strategies: random scaling  and random flipping. The random scaling strategy normalizes the  scores in the ISPP and the GT in the same proportion. The random flipping strategy converts the scores in the ISPP and GT between mean opinion score (MOS) and  differential mean opinion score (DMOS) with a probability. Through the two strategies, the PromptIQA must learn the assessment requirements through ISPPs, subsequently predicts specific scores. Experiments demonstrate the effectiveness of the two strategies in improving model performance and generalization.
Our contributions can be summarized as follows:

\begin{enumerate}
    \item We introduce a novel PromptIQA that can adapt to new assessment requirements through a small number of ISPs as prompts, eliminating the need for fine-tuning after the training is completed.
    \item We introduce two data augmentation strategies to make a strong correlation between the GT and the ISPP, which encourage PromptIQA to effectively learn more assessment requirements from prompts.
    \item Experimental results demonstrate that PromptIQA trained on mixed datasets through ISPPs exhibits outstanding performance and generalization. Furthermore, extensive experiments verify the PromptIQA can efficiently adapt to different assessment requirements through ISPPs. 
\end{enumerate}

\section{Related Works}
In the context of our work, we provide a brief review of related works on NR-IQA and mixed training for IQA.
\subsection{No-Reference Image Quality Assessment}

With the development of deep learning (DL), a growing number of DL-based NR-IQA models have been proposed, achieving significant performance and generalization. As the pioneers, Kang \etal \cite{kang2014convolutional} firstly design a convolutional neural networks (CNNs) for the IQA task to extract image features. Then they extend this work to a multi-task learning network  \cite{kang2015simultaneous}. However, as the DL-based networks become more complex, it is hard to train a reliable IQA models on the existing IQA datasets, which are in limited scales. Consequently, many models \cite{chen2024gmciqa,su2020blindly,yang2022maniqa} utilized models pretrained on image classification tasks, such as ResNet \cite{he2016deep} and vision transformer (ViT) \cite{dosovitskiy2020image}, as feature extractors. Nevertheless, recent studies \cite{zhu2020metaiqa, chen2022teacher} point out that these models pretrained on other tasks do not adapt well to the IQA task. Therefore, some works pretrain models on related pretext tasks, \emph{e.g.}, image restoration \cite{lin2018hallucinated, ma2021blind}, quality ranking \cite{liu2017rankiqa, ma2017dipiq}, and contrastive learning \cite{zhao2023quality, madhusudana2022image,shi2023transformer}. Recently, with the rapid development of large language models (LLM), some research attempts to combine natural language processing with the IQA task. For instance, Wang \etal\cite{wang2023exploring} and Saha \etal\cite{saha2023re} integrate textual information into the IQA. Wu \etal\cite{wu2023q} explore the possibility of integrating LLMs into low-level vision tasks. You \etal\cite{you2023depicting} propose an interpretable IQA model based on multi-modal LLMs.

\subsection{Mixed Training For Image Quality Assessment}
To break through the performance limitations of IQA models due to dataset scale constraints, one approach is to conduct mixed training across multiple IQA datasets. Although different IQA datasets contain distinct assessment requirements, standards, score ranges, types of labels and so on, mixing them for training and learning their common characteristics can effectively improve the performance and generalization of the IQA models. Hence, most of existing models try to learn from mixed datasets from two aspect: designing sophisticated networks and performing complex transformations on the datasets. For instance, Zhang \etal \cite{zhang2021uncertainty} sample image pairs from multiple datasets and take the probability of which image has a better perception calculated by human-rated scores and variances as labels. Finally, they train a unified NR-IQA model on these image pairs. Wang and Ma \cite{wang2021active} collect samples from datasets to fine-tune an unified model by active learning. Li \etal \cite{li2021unified} train a video quality assessment model with multiple datasets. Zhang \etal \cite{zhang2022continual} train the model to learn from a stream of IQA datasets by the continual learning strategy. Wang \etal \cite{wang2023Hierarchical} propose a novel multi-dataset training strategy by decomposing the mixed training into three hierarchical learning to learn the IQA task from easy to hard. Sun \etal \cite{sun2023blind} propose a StairIQA for training on mixed datasets by learning separate IQA regression heads for each dataset. While these heads for each dataset enable the execution of  specific evaluation tasks, such a design results in a  redundant  model structure  and limits the generalization on other unknown requirements.

Although existing models have improved IQA performance by addressing various aspects of the models and datasets, when there are changes in the assessment requirements in practical applications, these models are incapable of detecting changes. Consequently, the IQA models have to be retrained or fine-tuned on the corresponding datasets. However, theses datasets are time-consuming to obtain. In this paper, we propose an innovative framework for NR-IQA, which effectively transfers the model to the new assessment requirements with a small number of prompts without fine-tuning.

\section{The Proposed Method}

In this section, we firstly formalize the paradigms of typical IQA models and the PromptIQA.
Subsequently, we will provide a detailed exposition of the parts within the PromptIQA.

\subsection{Problem Formulation}

Given an image $I$, typical IQA models utilize an visual encoder $\mathcal{V}(\cdot)$ to extract visual features, followed by a regression model $\mathcal{R}(\cdot)$ to predict a quality score. This process can be represented as follows:
\begin{equation}
\label{eq:typical_iqa}
    S = \mathcal{R}(\mathcal{V}(I)).
\end{equation}
It indicates that once the training for $\mathcal{R}(\cdot)$ and $\mathcal{V}(\cdot)$ is completed, the assessment requirement inherent the model is unchanged during the inference stage. However, in different application scenarios, the assessment requirements vary. Consequently, it is necessary to re-train or fine-tune these models on the large scale of relevant datasets, which are not always available. On the contrary, the proposed PromptIQA can effectively adapt to new requirements by a few number of prompts without any fine-tuning, which can be formulated as follows:
\begin{equation}
\label{eq:promptIQA}
    S = \mathcal{R}(\mathcal{FM}_{IP}(\mathcal{V}(I), \mathcal{F}_{AP})),
\end{equation}
where  $\mathcal{F}_{AP}$ denotes the feature of the assessment prompt and $\mathcal{FM}_{IP}(\cdot)$ represents a fusion module. In this way, the trained $\mathcal{R}(\cdot)$ can be adapted to various assessment requirements by the $\mathcal{F}_{AP}$. In the experiments, we demonstrate that it is much more effective for adapting to the new assessment requirements. 
Next, we elaborate on how PromptIQA comprehends assessment requirements through prompts, thereby predicting score adaptively.

\subsection{Overview of the PromptIQA}
\label{sec:overview_promptiqa}

\begin{figure}[ht]
  \centering
  \vspace{-0.5cm}
  \includegraphics[width=\columnwidth]{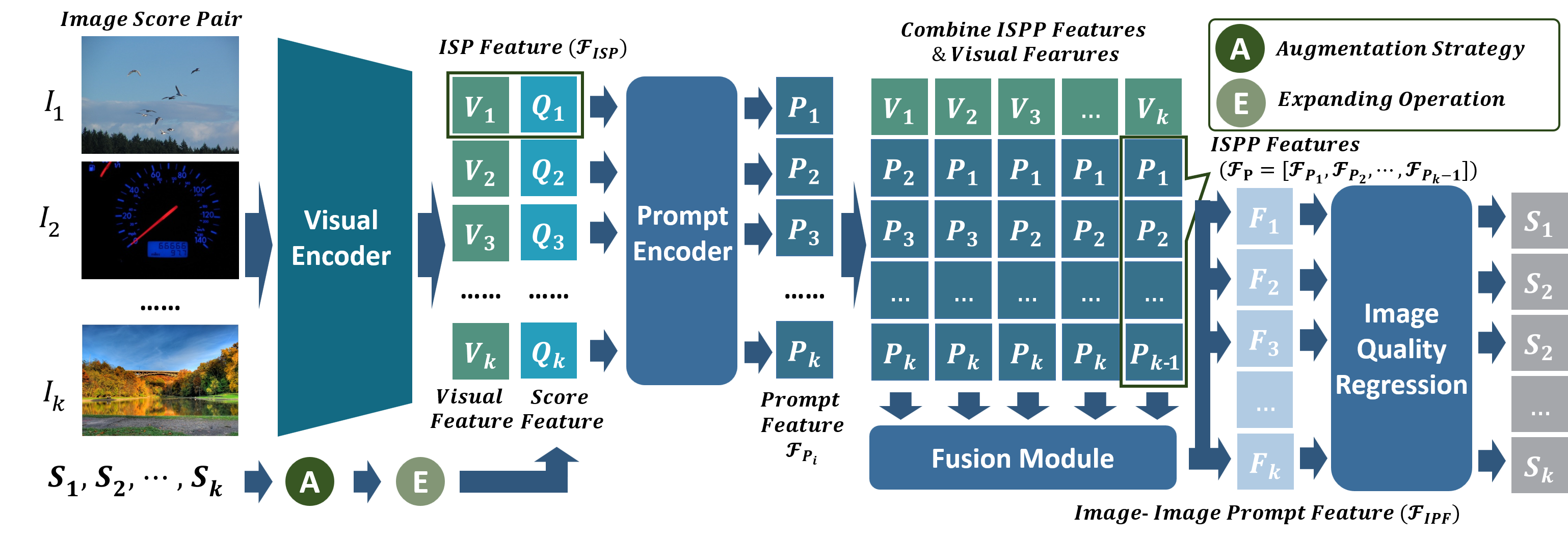}
  \caption{Overview of our proposed PromptIQA.}
  \vspace{-0.5cm}
  \label{fig:framework}
\end{figure}

As shown in \fig{fig:framework}, the PrompIQA consists of five parts: image-score pairs, a visual encoder, a prompt encoder, a fusion module and an image quality regression. We present details of the five parts in the following.

\noindent \textbf{A) Prompt Design.}  
Firstly, we introduce a novel concept: the Image-Score Pairs Prompt (ISPP), which consists of a sequence of image-score pairs (ISP) to reflect assessment requirements intuitively, such as assessment preferences and criteria, \etc.
Supposing $\mathcal{ISP}_i = (I_i, S_i)$ denotes the $i$-th ISP with the image $I_i$ and the corresponding quality score $S_i$. We collect $n$ ISPs to form an ISPP $\mathbf{P} = [ \mathcal{ISP}_1, \mathcal{ISP}_2, \cdots, \mathcal{ISP}_n ]$.

\noindent \textbf{B) Visual Encoder.}  
The primary objective of the visual encoder is to extract robust and effective visual representations related to the IQA task. In recent years, many IQA methods \cite{su2020blindly, chen2024gmciqa, ke2021musiq, qin2023data} have designed numerous efficient visual encoders. In this study, we adopt the Mean-opinion Network (MoNet) \cite{chen2024gmciqa} as the visual encoder. 

Then, we take the visual encoder $\mathcal{V}(\cdot)$ to extract the visual features $\mathcal{F}_{I_i} = \mathcal{V}(I_i) \in \mathbb{R}^{1\times N}$, where $1 \leq i \leq n$ and $N$ denotes the dimension, for each images in the ISPP. 
At the same time, we directly take the expanding operation $\mathcal{E}(\cdot)$ on the scores to match the dimensionality of the visual features to form the score feature $\mathcal{E}(S_i)$.
After that, we concatenate the visual feature $\mathcal{V}(I_i)$ and score feature $\mathcal{E}(S_i)$ together to form the $i$-th ISP feature, which can be formulated as follows:
\begin{equation}
    \mathcal{F}_{\mathcal{ISP}_i} = CAT(\mathcal{I}(I_i), \mathcal{E}(S_i)) \in \mathbb{R}^{2\times N},
\end{equation}
where $CAT(\cdot)$ represents the concatenation operation. 

\noindent \textbf{C) Prompt Encoder.} The role of this module is to explore the relationship between the visual feature and score feature in one ISP feature ($\mathcal{F}_{\mathcal{ISP}_i}$), which helps the PromptIQA comprehend why the image is labelled with this score. Although concatenation can combine the visual feature with score feature into a single feature, it is hard for the PromptIQA to explore the deep connections between the score and image from the feature. Therefore, we take 3 ViT blocks as the prompt encoder $\mathcal{P}(\cdot)$ to exploit the attention feature between visual features and score features in depth, denoted as $\mathcal{F}_f = \mathcal{P}(\mathcal{F}_{\mathcal{ISP}_i})\in \mathbb{R}^{2\times N}$. Then we take the average operation and a fully connected (FC) layer to reduce the dimension of $\mathcal{F}_f$ and gain the final prompt feature $\mathcal{F}_{P_i}\in\mathbb{R}^{1\times M}$, where $M$ represents the dimension. Finally, an ISPP feature $\mathcal{F}_{\mathbf{P}}=[\mathcal{F}_{P_1}, \mathcal{F}_{P_2}, \cdots, \mathcal{F}_{P_n}]$ that represents a high-dimensional projection for the assessment requirement is obtained.

\noindent \textbf{D) Fusion Module.}
There are two purposes of this module: I) to interact all the prompt features in the ISPP feature ($\mathcal{F}_{P_i}\in\mathcal{F}_{\mathbf{P}}$)  with each other, forming the assessment requirement feature $\mathcal{F}_{AC}$; II) to form the final image-prompt feature $\mathcal{F}_{IPF}$ by combining the visual features $\mathcal{V}(I)$ of the images to be evaluated with the assessment requirement features  $\mathcal{F}_{AC}$. Assuming that the ISPP feature consists of $n$ outputs of the prompt encoder, \ie $\mathcal{F}_\mathbf{P} = [\mathcal{F}_{P_1}, \mathcal{F}_{P_2}, \cdots, \mathcal{F}_{P_n}] \in \mathbb{R}^{n\times M}$, we firstly set 3 ViT blocks, denoted as an ISPP fusion module  $\mathcal{FM}_{ISPP}(\cdot)$, to make a profound interaction among the $n$ different prompt features to gain the assessment requirement feature $\mathcal{F}_{AC} = \mathcal{FM}_{ISPP}(\mathcal{F}_\mathbf{P})\in\mathbb{R}^{n\times M}$, thereby enabling the PromptIQA to further understand the assessment requirement. Then, we concatenate the visual features $\mathcal{V}(I)$ with the  $\mathcal{F}_{AC}$ and feed them into 8 ViT blocks, denoted as an image-prompt fusion module $\mathcal{FM}_{IP}(\cdot)$, to gain the final image-prompt feature $\mathcal{F}_{IPF} = \mathcal{FM}_{IP}(\mathcal{V}(I), \mathcal{F}_{AP})\in\mathbb{R}^{(n + 1)\times M}$.

\noindent \textbf{E) Image Quality Regression.} The purpose of the regression module $\mathcal{R}(\cdot)$ is to map the input features to quality scores. We take the first feature in image-prompt feature $\mathcal{F}_{IPF}\in\mathbb{R}^{(n + 1)\times M}$ as the input feature $\mathcal{F}_{Q}\in\mathbb{R}^{1\times M}$. The feature is processed through two FC layers that project the size from 128 to 64 and from 64 to 1. Finally, we obtain a predicted quality score $S = \mathcal{R}(\mathcal{F}_{Q})$.

\noindent \textbf{F) Augmentation Strategy.}
However, according to the aforementioned method, the PromptIQA may find a shortcut in the training process. The PromptIQA may ignore the ISPP and directly predict scores only depending on the input images. The optimization objective of the PromptIQA is generally formulated as minimizing the discrepancy between the predicted scores and the GT labels.
Due to the GT labels in one IQA dataset can only represent a single assessment requirement, the GTs remain unchanged even if the assessment prompt $\mathcal{F}_{AP}$ changes during the training stage. 
This undermines the role of prompts, thereby leading the PromptIQA to focus more on the input images for score prediction, rather than predicting scores through comprehending the prompts.

Therefore, it is necessary to enrich the assessment requirements in one datasets, forcing the PromptIQA to predict scores by learning the assessment requirements from the ISPP. We propose two data augmentation strategies, namely random scaling (RS) and random flipping (RF). The random scaling normalizes the scores in ISPP and GTs with a probability. The random flipping converts the scores in ISPP and GTs between MOS and DMOS with a probability. Assuming a list of $n + 1$ scores $\mathbf{S} = [S_1, S_2, \cdots, S_n, G]$, where $G$ represents the GT. The two strategies can be formulated as follows.
\begin{equation}
    \begin{split}
        f_{RS}(\mathbf{S}) &= \frac{1}{\max(\mathbf{S})}\cdot\mathbf{S} = \left[
        \frac{S_1}{\max(\mathbf{S})}, \frac{S_2}{\max(\mathbf{S})}, \cdots, \frac{S_n}{\max(\mathbf{S})} , \frac{G}{\max(\mathbf{S})} \right], \\
        f_{RF}(\mathbf{S}) &= \alpha - \mathbf{S} = \left[ \alpha - S_1, \alpha - S_2, \cdots, \alpha - S_n , \alpha - G\right],
    \end{split}
    \label{eq:rs_rf}
\end{equation}
where $\max(\cdot)$ can return the maximal value in the list and $\alpha$ is a constraint.
Through these two strategies, the PromptIQA must interpret assessment requirements via ISPP and then predicts targeted scores. Subsequent extensive experiments have demonstrated the effectiveness of these two strategies in improving model performance and generalization.

\section{Experiments}
\label{sec:experiments}

\subsection{Experiments Settings}
\noindent \textbf{A) Datasets.}
We collect a total of 12 datasets for 5 different IQA tasks, namely synthetic distortion nature IQA (SDN-IQA), authentic distortion nature IQA (ADN-IQA), face IQA (F-IQA), AI generated IQA (AIG-IQA) and underwater IQA (U-IQA). The detail information about these datasets is shown in \tbl{tab:detail_dataset}. Specifically, the images in the datasets designed for the SDN-IQA task are constructed by introducing synthetic distortions into reference images. Furthermore, the quality labels of images within these datasets are categorized into two types: MOS and DMOS. A higher MOS indicates better image quality; conversely, a higher DMOS signifies poorer image quality.

\begin{table}[]
\small
\vspace{-0.5cm}
\caption{Detail information about the 12 datasets.}
\label{tab:detail_dataset}
\resizebox{\columnwidth}{!}{
\begin{tabular}{c|c|c|c|c|c|c}
\hline
Dataset & Task & Ref. Image & Dist. Types & Image Number & Label Type & Range \\ \hline
LIVE~\cite{sheikh2006statistical} & \multirow{4}{*}{SDN-IQA} & \multicolumn{1}{c|}{29} & \multicolumn{1}{c|}{5} & \multicolumn{1}{c|}{779} & \multicolumn{1}{c|}{DMOS} & {[}1, 100{]} \\
CSIQ~\cite{larson2010most} &  & \multicolumn{1}{c|}{30} & \multicolumn{1}{c|}{6} & \multicolumn{1}{c|}{866} & \multicolumn{1}{c|}{DMOS} & {[}0, 1{]} \\
TID2013~\cite{ponomarenko2013color} &  & \multicolumn{1}{c|}{25} & \multicolumn{1}{c|}{24} & \multicolumn{1}{c|}{3,000} & \multicolumn{1}{c|}{MOS} & {[}0, 9{]} \\
Kadid-10k~\cite{kadid10k} &  & \multicolumn{1}{c|}{81} & \multicolumn{1}{c|}{25} & \multicolumn{1}{c|}{10,125} & \multicolumn{1}{c|}{MOS} & {[}1, 5{]} \\ \hline
BID~\cite{ciancio2010no} & \multirow{4}{*}{ADN-IQA} & \multicolumn{1}{c|}{-} & \multicolumn{1}{c|}{-} & \multicolumn{1}{c|}{586} & \multicolumn{1}{c|}{MOS} & {[}0, 5{]} \\
SPAQ~\cite{fang2020perceptual} &  & \multicolumn{1}{c|}{-} & \multicolumn{1}{c|}{-} & \multicolumn{1}{c|}{11,125} & \multicolumn{1}{c|}{MOS} & {[}0, 100{]} \\
LIVEC~\cite{ghadiyaram2015massive} &  & \multicolumn{1}{c|}{-} & \multicolumn{1}{c|}{-} & \multicolumn{1}{c|}{1,162} & \multicolumn{1}{c|}{MOS} & {[}1, 100{]} \\
KonIQ-10K~\cite{hosu2020koniq} &  & \multicolumn{1}{c|}{-} & \multicolumn{1}{c|}{-} & \multicolumn{1}{c|}{10,073} & \multicolumn{1}{c|}{MOS} & {[}0, 100{]} \\ \hline
GFIQA20k ~\cite{su2023going} & F-IQA & \multicolumn{1}{c|}{-} & \multicolumn{1}{c|}{-} & \multicolumn{1}{c|}{19,988} & \multicolumn{1}{c|}{MOS} & {[}0, 1{]} \\ \hline
AGIQA3k~\cite{li2023agiqa} & \multirow{2}{*}{AIG-IQA} & \multicolumn{1}{c|}{-} & \multicolumn{1}{c|}{-} & \multicolumn{1}{c|}{2,982} & \multicolumn{1}{c|}{MOS} & {[}0, 1{]} \\
AIGCIQA2023~\cite{wang2023aigciqa2023} &  & \multicolumn{1}{c|}{-} & \multicolumn{1}{c|}{-} & \multicolumn{1}{c|}{2,400} & \multicolumn{1}{c|}{MOS} & {[}0, 1{]} \\  \hline
UWIQA~\cite{yang2021reference} & U-IQA & \multicolumn{1}{c|}{-} & \multicolumn{1}{c|}{-} & \multicolumn{1}{c|}{890} & \multicolumn{1}{c|}{MOS} & {[}0, 1{]} \\ \hline
\end{tabular}
}
\vspace{-0.7cm}
\end{table}

\noindent \textbf{B) Evaluation Metrics.}
Spearman's rank order correlation coefficient (SROCC) and Pearson's linear correlation coefficient (PLCC) are used for evaluating the performance of IQA models. Both metrics range from -1 to 1, with higher absolute values indicating better performance of IQA models. 

\noindent \textbf{C) Implementation Details.}
Following the settings in  \cite{yang2022maniqa,qin2023data, su2020blindly}, we randomly divide each dataset into 80\% for training and 20\% for testing. This process is repeated 10 times, and the median of the 10 scores are reported as the final score. 
For the training and testing sets of each dataset, we independently apply the method of min-max normalization to normalize the label scores to the range of 0-1. Additionally, for testing, the ISPs are sampled from the training set. To ensure that there is no overlapping images between the training and the testing set on synthetic datasets, the datasets are divided according to the reference images. For other datasets, there are no reference images, so we divide the datasets according to the distorted images. 
We train PromptIQA directly on a combination of 12 datasets, without making any modifications to accommodate the different types of labels, such as MOS and DMOS.
For the SPAQ, the shortest side of an image is resized to 512 according to \cite{fang2020perceptual}. If not explicitly specified, the default number of ISPs in an ISPP is set to 10, the default ISP selection strategy is interval selection and the constraint $\alpha$ in \eq{eq:rs_rf} is set to 1. The probabilities of selecting the RS and RF strategies are 0.5 and 0.1 respectively.  We take $\mathcal{L}_1$ loss to diminish the discrepancy between predict scores and GTs. Then, we use the Adam optimizer with a learning rate of $1 \times 10^{-5}$ and a weight decay of $1 \times 10^{-5}$.  The learning rate is adjusted using the Cosine Annealing for every 50 epochs. We train our model for 100 epochs with a batch size of 66 on 6 NVIDIA 3090.

\subsection{Comparisons With State-Of-The-Arts}
\label{sec:compare_sota}
To demonstrate that PromptIQA can be efficiently trained on the combination of 12 datasets (referred to as the mixed dataset) and achieve a better performance, we conduct a comprehensive performance comparison of the PromptIQA with 21 state-of-the-art (SOTA) IQA methods on the 5 IQA tasks. Specifically, for the SDN-IQA and ADN-IQA tasks, our comparison encompasses 4 advanced DL-based IQA methods: HyperIQA~\cite{su2020blindly}, TOPIQ-NR~\cite{chen2023topiq}, MANIQA~\cite{yang2022maniqa}, and MoNet\cite{chen2024gmciqa}. Additionally, for the SDN-IQA task, we compare with another one traditional FR-IQA methods, SSIM~\cite{2004Image} and 2 DL-based FR-IQA method, namely LPIPS~\cite{2018TheLPIPS} and TOPIQ-FR~\cite{chen2023topiq}. For the F-IQA task, we take 4 task-specific F-IQA methods for comparison, namely SDD-FIQA~\cite{ou2021sdd}, IFQA~\cite{jo2023ifqa}, TOPIQ-Face~\cite{chen2023topiq}, and GPFIQA~\cite{su2023going}. For the AIG-IQA task, we include 4 task-specific IQA methods: Inception Score (IS)~\cite{NIPS2016_8a3363ab_IS}, NIQA~\cite{mittal2012making}, CLIPIQA~\cite{wang2022exploring}, and PSCR~\cite{yuan2023pscr}. In addition, for the U-IQA task, the comparison includes 4 task-specific IQA methods: FDUM~\cite{yang2021reference}, UCIQE~\cite{yang2015underwater}, URanker~\cite{guo2023uranker}, and UIQI~\cite{0UIQI}. Furthermore, we incorporate 2 mixed training IQA methods, namely UNIQUE~\cite{zhang2021uncertainty} and StairIQA~\cite{sun2023blind} for a holistic comparison. For task-specific models, training and evaluation are performed on the individual datasets, whereas for mixed training models, they are trained and evaluated across the mixed dataset. 

As shown in \tbl{tab:comparison_all}, when compared with the mixed training models, namely UNIQUE\footnote{Most datasets do not provide the standard deviation values for each MOS as required by the UNIQUE, resulting in incomplete experiment results.} and StairIQA, without complex processing of datasets and redundant model structure, PromptIQA exhibits powerful performance on the mixed datasets. 
Meanwhile, for the ADN-IQA, F-IQA, AIG-IQA, and U-IQA tasks, the PromptIQA outperforms nearly all the compared models. On the BID and LIVEC datasets, the PromptIQA surpasses the SOTA IQA models within a considerable margin of 1.53\% and 1.41\% in terms of SROCC, as well as 2.07\% and 1.21\% in terms of PLCC. However, on the SDN-IQA task, when compared to the DL-based FR-IQA method, \ie TOPIQ-FR, the PromptIQA show a slight lower performance. It is worth noting that the TOPIQ-FR takes the reference images as the input for training and testing to extract auxiliary features, which can greatly improve the performance. Nevertheless, our PromptIQA achieves better results to the variant, \ie TOPIQ-NR, on most datasets. For LIVE and CSIQ datasets, our model performs worse than most of the comparative methods. We speculate that there is a notable discrepancy in the distribution of DMOS labels within the LIVE and CSIQ datasets compared to the MOS labels in the other 10 datasets. However, the results in terms of SROCC and PLCC on LIVE (0.9359 and 0.9344) and CSIQ (0.9259 and 0.9392) are still acceptable. Nonetheless, the results demonstrate that the ISPPs can guide the PromptIQA to comprehend the different assessment requirements before judging the quality scores.

\begin{table}[ht]
\vspace{-0.5cm}
\caption{Comparison with state-of-the-art task specific models and mixed training models on 12 datasets for 5 IQA tasks\protect\footnotemark. The best and the second best scores are marked in black bold and blue, respectively.}
\renewcommand\arraystretch{0.99}
\setlength\tabcolsep{1.6pt}
\resizebox{\columnwidth}{!}{
\begin{tabular}{c|ccccccccccc}
\toprule
\hline
Task & \multicolumn{11}{c}{\textbf{\textit{Synthetic Distortion Nature IQA (SDN-IQA)}}} \\ \hline
 & \multicolumn{1}{c|}{Dataset} & \multicolumn{2}{c|}{LIVE} & \multicolumn{2}{c|}{CSIQ} & \multicolumn{2}{c|}{TID2013} & \multicolumn{2}{c|}{Kadid-10k} & \multicolumn{2}{c}{Average} \\ \cline{2-12} 
\multirow{-2}{*}{\begin{tabular}[c]{@{}c@{}}Training\\ Type\end{tabular}} & \multicolumn{1}{c|}{Method} & SROCC & \multicolumn{1}{c|}{PLCC} & SROCC & \multicolumn{1}{c|}{PLCC} & SROCC & \multicolumn{1}{c|}{PLCC} & SROCC & \multicolumn{1}{c|}{PLCC} & SROCC & PLCC \\ \hline
 & \multicolumn{1}{c|}{SSIM} & 0.8509 & \multicolumn{1}{c|}{0.7369} & 0.8437 & \multicolumn{1}{c|}{0.7685} & 0.6724 & \multicolumn{1}{c|}{0.7041} & 0.6196 & \multicolumn{1}{c|}{0.5750} & 0.7467 & 0.6961 \\
 & \multicolumn{1}{c|}{LPIPS} & 0.8690 & \multicolumn{1}{c|}{0.7672} & 0.9172 & \multicolumn{1}{c|}{0.9014} & 0.8468 & \multicolumn{1}{c|}{0.8137} & 0.8207 & \multicolumn{1}{c|}{0.7449} & 0.8634 & 0.8068 \\
 & \multicolumn{1}{c|}{TOPIQ-FR} & \textbf{0.9764} & \multicolumn{1}{c|}{{\color[HTML]{0070C0} \textbf{0.9639}}} & \textbf{0.9686} & \multicolumn{1}{c|}{\textbf{0.9739}} & \textbf{0.9527} & \multicolumn{1}{c|}{\textbf{0.9612}} & \textbf{0.9654} & \multicolumn{1}{c|}{\textbf{0.9671}} & \textbf{0.9658} & \textbf{0.9665} \\ \cline{2-12} 
 & \multicolumn{1}{c|}{HyperIQA} & 0.9499 & \multicolumn{1}{c|}{0.9522} & 0.9378 & \multicolumn{1}{c|}{0.9507} & 0.7459 & \multicolumn{1}{c|}{0.8027} & 0.8488 & \multicolumn{1}{c|}{0.8513} & 0.8706 & 0.8892 \\
 & \multicolumn{1}{c|}{TOPIQ-NR} & 0.9430 & \multicolumn{1}{c|}{0.9424} & 0.9076 & \multicolumn{1}{c|}{0.9247} & 0.8134 & \multicolumn{1}{c|}{0.8452} & 0.8770 & \multicolumn{1}{c|}{0.8754} & 0.8853 & 0.8969 \\
 & \multicolumn{1}{c|}{MANIQA} & 0.9562 & \multicolumn{1}{c|}{0.9599} & 0.9373 & \multicolumn{1}{c|}{0.9481} & 0.8859 & \multicolumn{1}{c|}{0.9047} & 0.9207 & \multicolumn{1}{c|}{0.9192} & 0.9250 & 0.9330 \\
\multirow{-7}{*}{\begin{tabular}[c]{@{}c@{}}Task\\ Specific\end{tabular}} & \multicolumn{1}{c|}{MoNet} & 0.9555 & \multicolumn{1}{c|}{0.9595} & {\color[HTML]{0070C0} \textbf{0.9549}} & \multicolumn{1}{c|}{{\color[HTML]{0070C0} \textbf{0.9613}}} & 0.8738 & \multicolumn{1}{c|}{0.8923} & 0.9208 & \multicolumn{1}{c|}{0.9226} & {\color[HTML]{0070C0} \textbf{0.9263}} & {\color[HTML]{0070C0} \textbf{0.9339}} \\ \hline
 & \multicolumn{1}{c|}{UNIQUE*} & {\color[HTML]{0070C0} \textbf{0.9690}} & \multicolumn{1}{c|}{\textbf{0.9680}} & 0.9020 & \multicolumn{1}{c|}{0.9270} & — & \multicolumn{1}{c|}{—} & 0.8780 & \multicolumn{1}{c|}{0.8760} & 0.9163 & 0.9237 \\
& \multicolumn{1}{c|}{StairIQA} & 0.9371 & \multicolumn{1}{c|}{0.9337} & 0.7679 & \multicolumn{1}{c|}{0.8430} & 0.6751 & \multicolumn{1}{c|}{0.7725} & 0.7853 & \multicolumn{1}{c|}{0.8051} & 0.7914 & 0.8386 \\
\multirow{-3}{*}{\begin{tabular}[c]{@{}c@{}}Mixed\\ Training\end{tabular}} & \multicolumn{1}{c|}{Ours} & 0.9359 & \multicolumn{1}{c|}{0.9344} & 0.9259 & \multicolumn{1}{c|}{0.9392} & {\color[HTML]{0070C0} \textbf{0.9030}} & \multicolumn{1}{c|}{{\color[HTML]{0070C0} \textbf{0.9223}}} & {\color[HTML]{0070C0} \textbf{0.9277}} & \multicolumn{1}{c|}{{\color[HTML]{0070C0} \textbf{0.9314}}} & 0.9231 & 0.9318 \\ \hline  \noalign{\vskip 4pt} \hline 
Task & \multicolumn{11}{c}{\textit{\textbf{Authentic Distortion Nature IQA (ADN-IQA)}}} \\ \hline
 & \multicolumn{1}{c|}{Dataset} & \multicolumn{2}{c|}{BID} & \multicolumn{2}{c|}{LIVEC} & \multicolumn{2}{c|}{SPAQ} & \multicolumn{2}{c|}{KonIQ10k} & \multicolumn{2}{c}{Average} \\ \cline{2-12} 
\multirow{-2}{*}{\begin{tabular}[c]{@{}c@{}}Training\\ Type\end{tabular}} & \multicolumn{1}{c|}{Method} & SROCC & \multicolumn{1}{c|}{PLCC} & SROCC & \multicolumn{1}{c|}{PLCC} & SROCC & \multicolumn{1}{c|}{PLCC} & SROCC & \multicolumn{1}{c|}{PLCC} & SROCC & PLCC \\ \hline
 & \multicolumn{1}{c|}{HyperIQA} & 0.8663 & \multicolumn{1}{c|}{0.8804} & 0.8507 & \multicolumn{1}{c|}{0.8718} & 0.9145 & \multicolumn{1}{c|}{0.9176} & 0.9105 & \multicolumn{1}{c|}{0.9216} & 0.8855 & 0.8979 \\
 & \multicolumn{1}{c|}{TOPIQ-NR} & 0.8179 & \multicolumn{1}{c|}{0.8218} & 0.8328 & \multicolumn{1}{c|}{0.8682} & 0.9137 & \multicolumn{1}{c|}{0.9165} & 0.9154 & \multicolumn{1}{c|}{0.9250} & 0.8699 & 0.8829 \\
 & \multicolumn{1}{c|}{MANIQA} & {\color[HTML]{0070C0} \textbf{0.9014}} & \multicolumn{1}{c|}{0.9104} & 0.8923 & \multicolumn{1}{c|}{0.9134} & 0.9218 & \multicolumn{1}{c|}{0.9250} & {\color[HTML]{0070C0} \textbf{0.9286}} & \multicolumn{1}{c|}{\textbf{0.9454}} & 0.9110 & 0.9235 \\
\multirow{-4}{*}{\begin{tabular}[c]{@{}c@{}}Task\\ Specific\end{tabular}} & \multicolumn{1}{c|}{MoNet} & 0.9012 & \multicolumn{1}{c|}{{\color[HTML]{0070C0} \textbf{0.9152}}} & {\color[HTML]{0070C0} \textbf{0.8998}} & \multicolumn{1}{c|}{{\color[HTML]{0070C0} \textbf{0.9169}}} & {\color[HTML]{0070C0} \textbf{0.9227}} & \multicolumn{1}{c|}{{\color[HTML]{0070C0} \textbf{0.9256}}} & 0.9284 & \multicolumn{1}{c|}{{\color[HTML]{0070C0} \textbf{0.9450}}} & {\color[HTML]{0070C0} \textbf{0.9130}} & {\color[HTML]{0070C0} \textbf{0.9257}} \\ \hline
 & \multicolumn{1}{c|}{UNIQUE*} & 0.8580 & \multicolumn{1}{c|}{0.8730} & 0.8540 & \multicolumn{1}{c|}{0.8900} & — & \multicolumn{1}{c|}{—} & 0.8960 & \multicolumn{1}{c|}{0.9010} & 0.8693 & 0.8880 \\
& \multicolumn{1}{c|}{StairIQA} & 0.7735 & \multicolumn{1}{c|}{0.7878} & 0.7795 & \multicolumn{1}{c|}{0.8549} & 0.9034 & \multicolumn{1}{c|}{0.9072} & 0.8651 & \multicolumn{1}{c|}{0.8955} & 0.8304 & 0.8614 \\
\multirow{-3}{*}{\begin{tabular}[c]{@{}c@{}}Mixed\\ Training\end{tabular}} & \multicolumn{1}{c|}{Ours} & \textbf{0.9152} & \multicolumn{1}{c|}{\textbf{0.9341}} & \textbf{0.9125} & \multicolumn{1}{c|}{\textbf{0.9280}} & \textbf{0.9228} & \multicolumn{1}{c|}{\textbf{0.9261}} & \textbf{0.9287} & \multicolumn{1}{c|}{0.9430} & \textbf{0.9198} & \textbf{0.9328} \\ \hline  \noalign{\vskip 4pt} \hline 
Task & \multicolumn{3}{c|}{\textbf{\textit{Face IQA (F-IQA)}}} & \multicolumn{5}{c|}{\textbf{\textit{AI Generate IQA (AIG-IQA)}}} & \multicolumn{3}{c}{\textbf{\textit{Underwater IQA(U-IQA)}}} \\ \hline
 & \multicolumn{1}{c|}{Dataset} & \multicolumn{2}{c|}{GFIQA20k} & \multicolumn{1}{c|}{Dataset} & \multicolumn{2}{c|}{AGIQA3k} & \multicolumn{2}{c|}{AIGCIQA2023} & \multicolumn{1}{c|}{Dataset} & \multicolumn{2}{c}{UWIQA} \\ \cline{2-12} 
\multirow{-2}{*}{\begin{tabular}[c]{@{}c@{}}Training\\ Type\end{tabular}} & \multicolumn{1}{c|}{Method} & SROCC & \multicolumn{1}{c|}{PLCC} & \multicolumn{1}{c|}{Method} & SROCC & \multicolumn{1}{c|}{PLCC} & SROCC & \multicolumn{1}{c|}{PLCC} & \multicolumn{1}{c|}{Method} & SROCC & PLCC \\ \hline
 & \multicolumn{1}{c|}{SDD-FIQA} & 0.6020 & \multicolumn{1}{c|}{0.6487} & \multicolumn{1}{c|}{IS} & 0.1839 & \multicolumn{1}{c|}{0.1857} & 0.2256 & \multicolumn{1}{c|}{0.2336} & \multicolumn{1}{c|}{FDUM} & 0.6940 & 0.6685 \\
 & \multicolumn{1}{c|}{IFQA} & 0.6969 & \multicolumn{1}{c|}{0.7221} & \multicolumn{1}{c|}{NIQE*} & 0.5623 & \multicolumn{1}{c|}{0.5171} & 0.5060 & \multicolumn{1}{c|}{0.5218} & \multicolumn{1}{c|}{UCIQE*} & 0.6271 & 0.6261 \\
 & \multicolumn{1}{c|}{TOPIQ-Face} & {\color[HTML]{0070C0} \textbf{0.9664}} & \multicolumn{1}{c|}{{\color[HTML]{0070C0} \textbf{0.9668}}} & \multicolumn{1}{c|}{CLIPIQA*} & 0.8426 & \multicolumn{1}{c|}{0.8053} & — & \multicolumn{1}{c|}{—} & \multicolumn{1}{c|}{URanker} & 0.6744 & 0.6626 \\
\multirow{-4}{*}{\begin{tabular}[c]{@{}c@{}}Task\\ Specific\end{tabular}} & \multicolumn{1}{c|}{GPFIQA*} & 0.9643 & \multicolumn{1}{c|}{0.9652} & \multicolumn{1}{c|}{PSCR*} & {\color[HTML]{0070C0} \textbf{0.8498}} & \multicolumn{1}{c|}{\textbf{0.9059}} & {\color[HTML]{0070C0} \textbf{0.8371}} & \multicolumn{1}{c|}{{\color[HTML]{0070C0} \textbf{0.8588}}} & \multicolumn{1}{c|}{UIQI*} & {\color[HTML]{0070C0} \textbf{0.7423}} & {\color[HTML]{0070C0} \textbf{0.7412}} \\ \hline
 & \multicolumn{1}{c|}{UNIQUE*} & — & \multicolumn{1}{c|}{—} & \multicolumn{1}{c|}{UNIQUE*} & — & \multicolumn{1}{c|}{—} & — & \multicolumn{1}{c|}{—} & \multicolumn{1}{c|}{UNIQUE*} & — & — \\
& \multicolumn{1}{c|}{StairIQA} & 0.9374 & \multicolumn{1}{c|}{0.9350} & \multicolumn{1}{c|}{StairIQA} & 0.7553 & \multicolumn{1}{c|}{0.8303} & 0.7546 & \multicolumn{1}{c|}{0.7821} & \multicolumn{1}{c|}{StairIQA} & 0.7224 & 0.7270 \\
\multirow{-3}{*}{\begin{tabular}[c]{@{}c@{}}Mixed\\ Training\end{tabular}} & \multicolumn{1}{c|}{Ours} & \textbf{0.9698} & \multicolumn{1}{c|}{\textbf{0.9702}} & \multicolumn{1}{c|}{Ours} & \textbf{0.8509} & \multicolumn{1}{c|}{{\color[HTML]{0070C0} \textbf{0.9013}}} & \textbf{0.8508} & \multicolumn{1}{c|}{\textbf{0.8743}} & \multicolumn{1}{c|}{Ours} & \textbf{0.8766} & \textbf{0.8839} \\ \hline \bottomrule
\end{tabular}
}
\label{tab:comparison_all}
\end{table}
\footnotetext{* indicates that the results are referred from the original paper.}

\subsection{Generalization Comparison on Other Assessment Requirements}
\label{sec:generalization}

To evaluate the generalization of the PromptIQA for unknown assessment requirements, we employ 3 FR-IQA models, namely SSIM\cite{2004Image}, FSIM\cite{2011FSIM} and LPIPS\cite{2018TheLPIPS} to simulate 3 different assessment requirements. These models are applied to 2 synthetic distortion datasets (TID2013~\cite{ponomarenko2013color} and Kadid-10k~\cite{kadid10k}) to generate new quality labels, forming 6 reconstituted datasets. In these datasets, the labels generated by SSIM and FSIM are similar to MOS, indicating that a higher score reflects better quality.  By contrast, those generated by LPIPS resemble DMOS, which indicates a   higher score denotes poorer quality. In this experiment, 10 ISPs are extracted from the training sets of these 6 reconstituted datasets respectively, forming 6 types of ISPPs to indicate the assessment requirements for the PromptIQA.

We compare the PromptIQA with MANIQA\cite{yang2022maniqa} and MoNet\cite{chen2024gmciqa} in the following training modes:
I) \textbf{Specific dataset training (SDT):} The model is trained on TID2013 and Kadid-10k with the original labels respectively;
II) \textbf{Mixed dataset training (MDT):} The model is trained on the mixed dataset as described in Sec.\ref{sec:compare_sota}. Since the PromptIQA takes the ISPP as input to extract auxiliary features during the inference stage, to ensure a fair comparison, we designed another two fine-tuning modes for the comparison models.
III) \textbf{Specific dataset training with fine-tuning (SDT \& FT):} On the basis of models trained on specific datasets, the model is fine-tuned on the corresponding selected 10 ISPs;
IV) \textbf{Mixed dataset training with fine-tuning (MDT \& FT):} On the basis of models that trained on the mixed dataset, fine-tuning is conducted on the corresponding 10 selected ISPs. After training, all these models are evaluated on the test sets of the 6 reconstituted datasets.

As shown in \tbl{tab:Generalization_Comparison_On_Other_Assessment_Metrics}, without any fine-tuning, PromptIQA-MDT demonstrates excellent generalization with just 10 ISPs as prompt. In contrast, comparative models trained on specific datasets (MANIQA-SDT and MoNet-SDT) and the mixed dataset (MANIQA-MDT and MoNet-MDT) struggle to adapt to new assessment requirements and exhibit low generalization. The reason for this is that typical IQA models struggle to detect changes in requirements after training, which leads them to predict scores only depending on the requirement learning from the training datasets. This impacts their generalization and the results also indicate that there are significant differences among various assessment requirements. Furthermore, since there is not guidance for MANIQA-MDT and MoNet-MDT to distinguish different datasets in mixed training, the results are even worse than those trained on the specific datasets. Nevertheless, benefiting from the assistance of ISPP in mixed training, PromptIQA-MDT is not  confused by the discrepancy among various datasets.

When compared to MANIQA-SDT\&FT and MoNet-SDT\&FT, which are additionally fine-tuned, they still show a limited generalization. This is because such a small number of data is not sufficient enough for these models to comprehend the new assessment requirements. Since the PromptIQA-SDT is only trained on the single dataset, it can not also adapt well to other assessment requirements. But it still shows better generalization on most datasets than MANIQA-SDT\&FT and MoNet-SDT\&FT.
Thanks to training on the mixed dataset, PromptIQA-MDT has learned multiple assessment requirements. This makes the PromptIQA can easily adapt to new requirements with only a few ISPs as prompts without fine-tuning. 
Moreover, when evaluating on the datasets created by the LPIPS, we observe that the results for MANIQA and MoNet are negative. This is due to the disparity between LPIPS (DMOS) and the training labels (MOS). Nevertheless, PromptIQA can recognize this discrepancy and make accurate and reasonable predictions. 

\begin{table}[htbp]
\caption{Generalization comparison on other assessment requirements in terms of PLCC. The best and second best scores are marked in black bold and blue, respectively.}
\label{tab:Generalization_Comparison_On_Other_Assessment_Metrics}
\resizebox{\columnwidth}{!}{
\begin{tabular}{ccc|ccc|ccc}
\hline
\multicolumn{3}{c|}{Dataset} & \multicolumn{3}{c|}{TID2013(MOS)} & \multicolumn{3}{c}{Kadid-10k(MOS)} \\ \hline
\multicolumn{1}{c|}{Models} & Training Mode & Test Type & SSIM(MOS) & FSIM(MOS) & LPIPS(DMOS) & SSIM(MOS) & FSIM(MOS) & LPIPS(DMOS) \\ \hline
\multicolumn{1}{c|}{} & SDT & Zero-shot & 0.5391 & 0.8245 & -0.7486 & 0.5553 & 0.6902 & -0.7159 \\
\multicolumn{1}{c|}{} & SDT \& FT & Few-shot & 0.5583 & 0.8422 & {\color[HTML]{0070C0} \textbf{-0.7668}} & 0.5558 & 0.6959 & -0.7156 \\ \cline{2-9} 
\multicolumn{1}{c|}{} & MDT & Zero-shot & 0.4427 & 0.6869 & -0.6238 & 0.5644 & 0.6872 & -0.7477 \\
\multicolumn{1}{c|}{\multirow{-4}{*}{MANIQA}} & MDT \& FT & Few-shot & 0.4507 & 0.6925 & -0.6202 & 0.5652 & 0.6884 & -0.7474 \\ \hline
\multicolumn{1}{c|}{} & SDT & Zero-shot & 0.5322 & 0.8143 & -0.7384 & 0.5704 & 0.6886 & -0.7372 \\
\multicolumn{1}{c|}{} & SDT \& FT & Few-shot & 0.5473 & 0.8380 & -0.7478 & 0.5708 & 0.6983 & -0.7301 \\ \cline{2-9} 
\multicolumn{1}{c|}{} & MDT & Zero-shot & 0.4346 & 0.5433 & -0.5215 & 0.5689 & 0.6870 & -0.7426 \\
\multicolumn{1}{c|}{\multirow{-4}{*}{MoNet}} & MDT \& FT & Few-shot & 0.4413 & 0.5553 & -0.5034 & {\color[HTML]{0070C0} \textbf{0.5716}} & 0.6945 & -0.7399 \\ \hline
\multicolumn{1}{c|}{} & SDT & Few-shot & {\color[HTML]{0070C0} \textbf{0.5594}} & {\color[HTML]{0070C0} \textbf{0.8579}} & 0.7511 & 0.5615 & {\color[HTML]{0070C0} \textbf{0.7038}} & {\color[HTML]{0070C0} \textbf{0.7567}} \\
\multicolumn{1}{c|}{\multirow{-2}{*}{PromptIQA}} & MDT & Few-shot & \textbf{0.5992} & \textbf{0.8802} & \textbf{0.8064} & \textbf{0.5717} & \textbf{0.7196} & \textbf{0.7615} \\ \hline
\end{tabular}
}
\vspace{-0.8cm}
\end{table}

\subsection{Prompt Discussion}

\noindent\textbf{A) Discussion about the ISP Selection Strategies.}
In this part, we explore the effect of two ISP selection strategies, namely random sampling and interval sampling on the performance of PromptIQA. Random sampling involves selecting ISPs randomly from the training set. Interval sampling entails ranking the training set by score labels and subsequently selecting ISPs at uniform intervals. As we can see from the \tbl{tab:prompt_selection_mode}, ISPs chosen by interval sampling provide better guidance to the PromptIQA, because the interval sampling align more with the distribution of the original dataset. However, even though ISPs  sampled randomly can lead to a decrease in performance, the SROCC and PLCC still remain at high levels with a relative low standard deviation (STD), which demonstrates the robustness of PromptIQA to ISPs.

\begin{table}[ht]
\caption{The impact of different ISP selection strategies on the performance of the PromptIQA. The best scores are marked in black bold.}
\label{tab:prompt_selection_mode}
\resizebox{\columnwidth}{!}{
\begin{tabular}{c|cc|cc|cc|cc|cc|cc}
\hline
Dataset & \multicolumn{2}{c|}{Kadid-10k} & \multicolumn{2}{c|}{SPAQ} & \multicolumn{2}{c|}{KonIQ10k} & \multicolumn{2}{c|}{GFIQA20k} & \multicolumn{2}{c|}{AIGCIQA2023} & \multicolumn{2}{c}{UWIQA} \\ \hline
Strategy & SROCC & PLCC & SROCC & PLCC & SROCC & PLCC & SROCC & PLCC & SROCC & PLCC & SROCC & PLCC \\ \hline
Random & 0.9179 & 0.9204 & 0.9194 & 0.9207 & 0.9201 & 0.9345 & 0.9622 & 0.9631 & 0.8423 & 0.8654 & \textbf{0.8773} & \textbf{0.8843} \\
STD & 0.0169 & 0.0151 & 0.0035 & 0.0035 & 0.0156 & 0.0101 & 0.0038 & 0.0032 & 0.0142 & 0.0137 & 0.0181 & 0.0179 \\ \hline
Interval & \textbf{0.9277} & \textbf{0.9314} & \textbf{0.9228} & \textbf{0.9261} & \textbf{0.9287} & \textbf{0.9430} & \textbf{0.9698} & \textbf{0.9702} & \textbf{0.8508} & \textbf{0.8743} & 0.8766 & 0.8839 \\ \hline
\end{tabular}
}
\end{table}

\begin{figure}[ht]
\vspace{-0.5cm}
    \centering
    \includegraphics[width=\columnwidth]{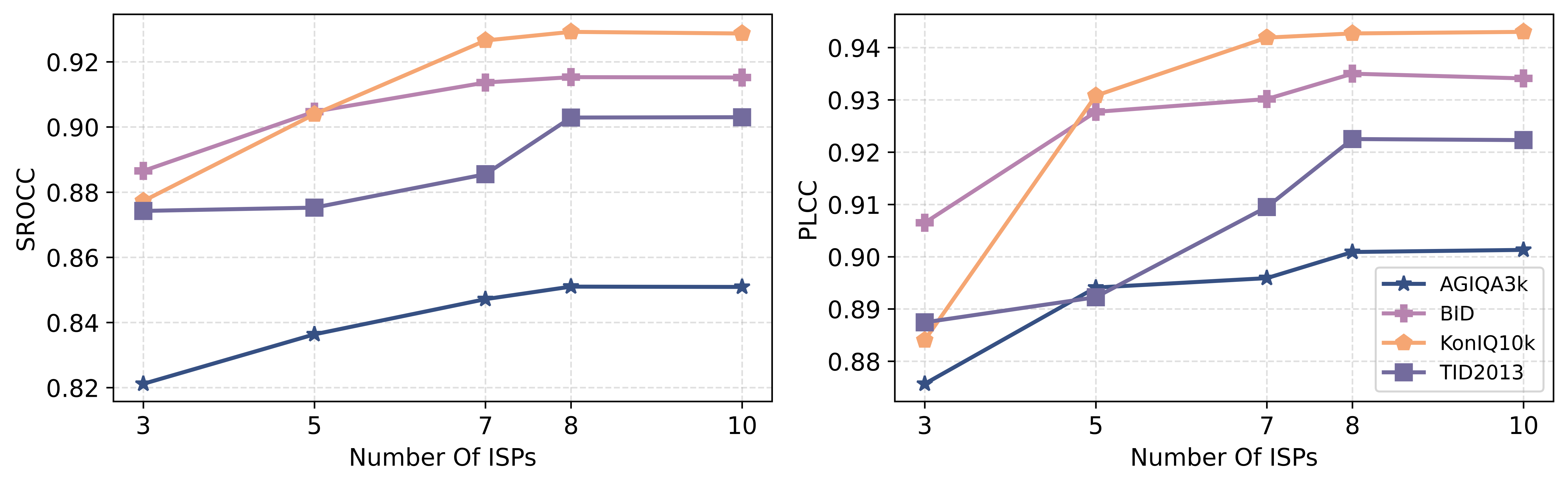}
    \caption{The impact of the number of ISPs in 
    an ISPP on PromptIQA performance.}
    \label{fig:size_of_ispp}
    \vspace{-0.3cm}
\end{figure}

\noindent\textbf{B) Discussion about the Size of ISPP.}
To investigate the effect of the number of ISPs in one ISPP on the performance of the PromptIQA, we set the number of ISPs to 3, 5, 7, 8 and 10. The results on the LIVEC, BID and SPAQ datasets are shown in \fig{fig:size_of_ispp}. As the number of ISPs increases, the performance of the PromptIQA increases intuitively. This is because the more ISPs are used as the prompt, the more guiding information is transmitted to the PromptIQA, therefore leading to more effective learning on the assessment requirement. 
 
\noindent\textbf{C) The Effect of ISP Prompts.}
 In this part, we employ two strategies, namely random and inversion, to quantify the effect of the ISPP on the PromptIQA. The strategy of random refers to the replace images or scores in the ISPP by generated random numbers, while the inversion method employs the transformation $x_i^\prime=1-x_i$ to reverse the original label $x_i$ in the ISPP. As shown in the \tbl{tab:Quantified_The_Effect_Of_Prompts}, randomizing any parts in ISPP significantly reduces the performance of PromptIQA. This decline in performance is attributed to the meaningless assessment requirements in the randomized ISPPs, making PromptIQA fail to learn. Additionally, since the label types between inverted scores in the ISPP and the original labels are opposite, namely higher scores indicating better image quality and higher scores indicating worse image quality, it leads to negative values in terms of SROCC and PLCC. Nonetheless, such inversion does not affect the performance of PromptIQA, indicating its robustness to varied prompts. 

\begin{table}[ht]
\vspace{-0.5cm}
\caption{The effect of ISP prompts. The check marked part in the ISPP denotes to apply the corresponding checked strategy, while unmarked part rely on original data. The best scores are marked in black bold.}
\centering
\label{tab:Quantified_The_Effect_Of_Prompts}
\resizebox{\columnwidth}{!}{
\begin{tabular}{cccc|cc|cc|cc|cc|cc}
\hline
\multicolumn{2}{c}{Strategy} & \multicolumn{2}{c|}{ISPP} & \multicolumn{2}{c|}{TID2013} & \multicolumn{2}{c|}{KonIQ10k} & \multicolumn{2}{c|}{GFIQA20k} & \multicolumn{2}{c|}{AGIQA3k} & \multicolumn{2}{c}{UWIQA} \\ \hline
Random & Inversion & Image & Score & SROCC & PLCC & SROCC & PLCC & SROCC & PLCC & SROCC & PLCC & SROCC & PLCC \\ \hline
\checkmark &  & \checkmark &  & 0.1307 & 0.0809 & 0.3624 & 0.3826 & 0.0645 & 0.0909 & -0.0786 & -0.0523 & 0.5582 & 0.7313 \\
\checkmark &  &  & \checkmark & -0.1181 & -0.2152 & 0.0299 & -0.0098 & -0.1346 & -0.2682 & 0.2792 & 0.3284 & 0.4939 & 0.4939 \\
\checkmark &  & \checkmark & \checkmark & 0.1287 & 0.0073 & 0.4374 & 0.2741 & 0.6143 & 0.6938 & -0.0617 & 0.0224 & 0.1162 & -0.2466 \\
 & \checkmark &  & \checkmark & -0.9024 & -0.9212 & -0.9264 & -0.9425 & \textbf{-0.9698} & -0.9690 & -0.8481 & \textbf{-0.8743} & \textbf{-0.8769} & -0.8765 \\ \hline
\multicolumn{4}{c|}{Standard ISPP} & \textbf{0.9030} & \textbf{0.9223} & \textbf{0.9287} & \textbf{0.9430} & \textbf{0.9698} & \textbf{0.9702} & \textbf{0.8508} & \textbf{0.8743} & 0.8766 & \textbf{0.8839} \\ \hline
\end{tabular}
}
\end{table}

\subsection{Ablation Study}

To validate the effectiveness of the proposed key components, we train five variants of the PromptIQA:
I) w/o mixed training, which is directly trained and tested on the single dataset;
II) w/o prompt, which removes all components related to prompt, namely the ISPP, prompt encoder and fusion module; 
III) w/o random scaling and IV) w/o random flipping, which remove the random scaling and random flipping strategies defined in Eq. (\ref{eq:rs_rf}), respectively; 
V) w/o random scaling \& flipping, which removes the two strategies simultaneously. Except the w/o mixed training variant, other variants are trained and tested on the mixed dataset as described in the Sec.\ref{sec:compare_sota}. Then the generalization comparison experiments, detailed in Sec.\ref{sec:generalization},  are conducted on all these trained variants.

The results from \tbl{tab:ablation} reveal that the removal of any components degrades the model's performance and generalization. We can see that the performance and generalization of the full model surpass those of the w/o mixed training variant. This is because full model can effectively learn the generic quality features and various assessment requirements from the mixed dataset.
Meanwhile, removing the prompt leads to a significant decrease in performance during mixed training and generalization test, which indicates that the prompt can help the PromptIQA distinct the different assessment requirements in the mixed datasets and adapt easily to the new requirements. Moreover, although the variants of eliminating the random scaling and flipping strategies demonstrate  close performance to the full model in the mixed training, they show a significantly lower generalization. This suggests that without data augmentation strategies, the model can find a shortcut to make the prompt ineffectiveness, consequently leading to degraded generalization. 

\begin{table}[ht]
\vspace{-0.5cm}
\caption{Ablation studies on the critical components of the PromptIQA in terms of PLCC. The best scores are marked in black bold.}
\label{tab:ablation}
\centering
\resizebox{\columnwidth}{!}{
\begin{tabular}{c|cccc|cccc}
\hline
\multirow{3}{*}{Ablation Study Settings} & \multicolumn{4}{c|}{\multirow{2}{*}{Testing Set On The Mixed Dataset}} & \multicolumn{4}{c}{Generalization Test} \\ \cline{6-9} 
 & \multicolumn{4}{c|}{} & \multicolumn{2}{c|}{TID2013(MOS)} & \multicolumn{2}{c}{Kadid-10k} \\ \cline{2-9} 
 & TID2013 & SPAQ & GFIQA20k & UWIQA & FSIM(MOS) & \multicolumn{1}{c|}{LPIPS(DMOS)} & SSIM(MOS) & LPIPS(DMOS) \\ \hline
w/o mixed training & 0.8849  &  0.9228 &	0.9696  &	0.8781   & 0.8579 &  \multicolumn{1}{c|}{0.6511} & 0.5615 & 0.7267  \\
w/o prompt & 0.8929 & 0.9220 & 0.9665 & 0.8602 & 0.7851 & \multicolumn{1}{c|}{-0.7712} & 0.5632 & -0.7397 \\ \hline
w/o random scale & 0.9218 & 0.9252 & 0.9683 & 0.8766 & 0.8614 & \multicolumn{1}{c|}{0.7797} & 0.5692 & 0.7458 \\
w/o random flipping & 0.9080 & 0.9245 & 0.9691 & 0.8754 & 0.8646 & \multicolumn{1}{c|}{0.6538} & 0.5612 & 0.5711 \\
w/o random scale \& flipping & 0.9028 & 0.9248 & 0.9681 & 0.8791 & 0.8370 & \multicolumn{1}{c|}{0.6243} & 0.5565 & 0.4960 \\ \hline
Full Model & \textbf{0.9223} & \textbf{0.9261} & \textbf{0.9702} & \textbf{0.8839} & \textbf{0.8802} & \multicolumn{1}{c|}{\textbf{0.8064}} & \textbf{0.5717} & \textbf{0.7615} \\ \hline
\end{tabular}
}
\vspace{-0.5cm}
\end{table}

\section{Conclusion}

In this paper, we propose a PromptIQA, which can effectively adapt to the new assessment requirements with a few number of image-score pairs as the prompt. Additionally, we analyze that directly training PromptIQA with datasets leads the model to find a shortcut, ignoring the changes in prompts. This makes the prompts ineffective and reduces model generalization. To address this issue, we propose two data augmentation strategies, random scaling and random flipping, that dynamically adjust the relationship between GT and prompt during the training process, thereby forcing the model to learn assessment requirements through the prompt. Moreover, experiments demonstrate that the PromptIQA trained with the mixed dataset shows an excellent performance, and can easily adapt to new assessment requirements. The ablation studies verify the effectiveness of proposed key components.

\bibliographystyle{splncs04}
\bibliography{main}
\end{document}